\pdfoutput=1

\documentclass[11pt]{article}

\usepackage[]{acl}

\usepackage{times}
\usepackage{latexsym}

\usepackage[T1]{fontenc}

\usepackage[utf8]{inputenc}

\usepackage{microtype}

\usepackage{times}  
\usepackage{graphicx} 
\usepackage{natbib}  
\usepackage{diagbox}
\usepackage{caption} 
\usepackage[utf8]{inputenc}
\usepackage[T1]{fontenc}
\usepackage[english]{babel}
\usepackage{amsfonts}
\usepackage{nicefrac}
\usepackage{microtype}
\usepackage{amsmath}
\usepackage{amssymb}
\usepackage{mathtools}
\usepackage{subcaption}
\usepackage{multirow}
\usepackage{booktabs}
\usepackage{ragged2e}
\usepackage{tikz}
\usepackage{stackengine}
\usepackage{etoolbox}
\usepackage{xspace}
\usepackage{xpatch}
\usepackage{cuted}
\usepackage{enumerate}
\usepackage{xstring}
\usepackage{setspace}
\usepackage{tabularx}
\usepackage{makecell}
\usepackage{changepage}
\usepackage{enumitem}
\usepackage{cuted}
\usepackage{titlesec}
\usepackage{cancel}
\usepackage{eqparbox}
\usepackage{bibentry}
\usepackage{amsmath}

\usepackage{algpseudocode}
\usepackage[hang,flushmargin]{footmisc}
\usepackage[capitalise,noabbrev,nameinlink]{cleveref}
\usepackage[useregional=numeric]{datetime2}
\usepackage[linesnumbered,ruled,noend,noline]{algorithm2e}
\usepackage{adjustbox}

\SetKwInput{KwModel}{Model}
\SetKwInput{KwParameter}{Parameter}
\SetKw{KwBy}{by}
\SetKw{KwIn}{in}
\SetKwInput{KwMetric}{Metric}

\usepackage[switch,mathlines]{lineno}
\usepackage{lipsum}
\usepackage{etoolbox}
\usepackage{longtable}

\makeatletter
\def\@opargbegintheorem#1#2#3{\trivlist
   \item[]{\textbf{#1\ #2} \ (#3)\textbf{.}\ } \itshape}
\makeatother

\graphicspath{{figures/}}

\title{    
    Query Expansion Using Contextual Clue Sampling with \\ Language Models 
}

\author{
    Linqing Liu$^{\dagger{}}$ \hspace{0.15cm}
    Minghan Li$^{\ddagger{}}$ \hspace{0.15cm} 
    Jimmy Lin$^{\ddagger{}}$ \hspace{0.15cm}
    Sebastian Riedel$^{\dagger{}}$ \hspace{0.15cm}
    Pontus Stenetorp$^{\dagger{}}$ \hspace{0.15cm} \\
    $^{\dagger{}}$University College London \hspace{0.3cm} $^{\ddagger{}}$University of Waterloo\\
    \texttt{\{linqing.liu,s.riedel,p.stenetorp\}@cs.ucl.ac.uk}  \\
    \texttt{\{m692li,jimmylin\}@uwaterloo.ca} \\
}

\begin{document}
\maketitle

\newcommand{\anscontext}{contextual clues}
\newcommand{\anscontextsingle}{contextual clue}
\newcommand{\filteredcontext}{prototypes}

\begin{abstract}
Query expansion is an effective approach for mitigating vocabulary mismatch between queries and documents in information retrieval.
%
One recent line of research uses language models to generate query-related contexts for expansion. 
Along this line, we argue that expansion terms from these contexts should balance two key aspects: diversity and relevance.
The obvious way to increase diversity is to sample multiple contexts from the language model.
However, this comes at the cost of relevance, because there is a well-known tendency of models to hallucinate incorrect or irrelevant contexts.
To balance these two considerations, we propose a combination of an effective filtering strategy and fusion of the retrieved documents based on the generation probability of each context.
Our lexical matching based approach achieves a similar top-5/top-20 retrieval accuracy and higher top-100 accuracy compared with the well-established dense retrieval model DPR, while reducing the index size by more than 96\%.
For end-to-end QA, the reader model also benefits from our method and achieves the highest Exact-Match score against several competitive baselines.
\end{abstract}

\section{Introduction}
Despite the advent of dense retrieval approaches based on semantic matching for open-domain question answering such as DPR~\cite{karpukhin2020dense}, approaches based on lexical matching~(e.g., BM25) remain important due to their space-efficiency and can serve as input to hybrid methods~\cite{gao2021coil,Formal2021SPLADESL,Lin2021AFB}.
%
%

%
A core challenge for lexical retrieval is the vocabulary mismatch between the query and documents.
Query expansion techniques dating back over half a century have proven effective in overcoming this issue~\cite{salton1971smart}.
The expansion terms are traditionally precomputed from relevant corpora using pseudo-relevance feedback techniques~\cite{salton1971smart, robertson1976relevance, abdul2004umass}.
In recent work, GAR~\cite{mao2021generation} explored removing the query expansion's reliance on an external corpus and instead used a large language model to generate a context.

We argue that expansion needs to balance two key factors:
(1) Diversity: Given the question, there can be multiple different reasoning paths~(referred to as \textit{\anscontext{}}) to reach the correct answer.
(2) Relevance: Simply relying on a single generated context increases the risk of query drift, as the generated context could be semantically irrelevant or contain factual errors~\cite{schutze2008introduction}.
However, simply generating multiple contexts is prone to the \textit{hallucination} problem -- they can be unfaithful to the input or include false information~\cite{tian2019sticking, maynez2020faithfulness, dziri2021neural}.
Thus, in this work, we wish to explore the question: How can we best generate a sufficiently rich set of \anscontext{} to answer a query?

\begin{figure}
\centering
\includegraphics[width=0.9\linewidth]{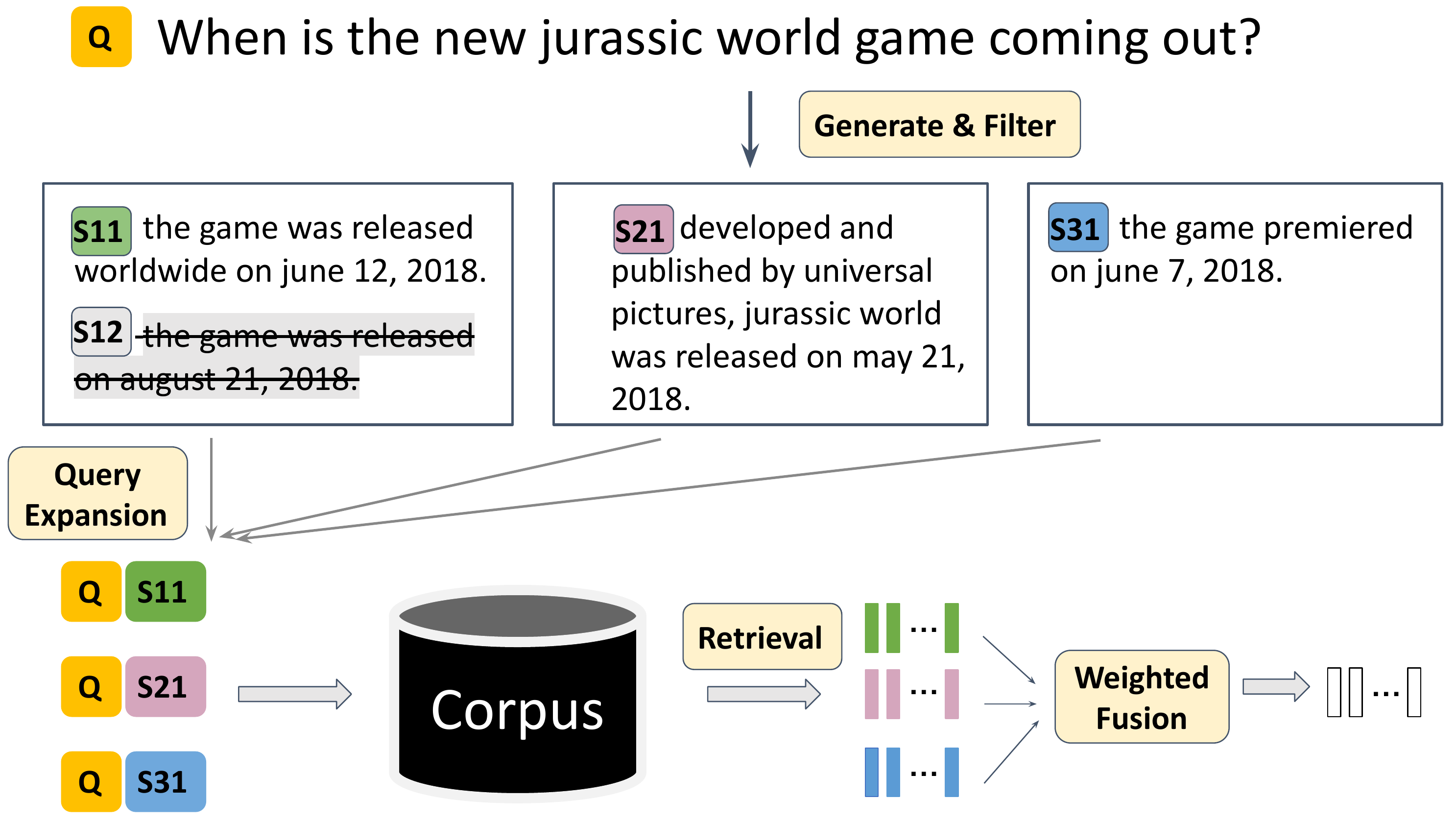}
\caption{
Given the query, we first sample various contextual clues from the language model.
Lexically similar outputs are grouped together and we only keep the output with the highest generation probability~(\textit{Filtering}).
We then perform retrieval for each augmented query individually.
All the retrieved documents are then ranked together in the \textit{fusion} step.  
}
\label{fig:intro}
\end{figure}

Our proposed solution~(Figure~\ref{fig:intro}) overcomes these problems with two simple and efficient steps: Filtering and fusion.
%
%
After sampling top-k outputs from the decoder of the fine-tuned language model, we first cluster these generated \anscontext{} based on their lexical distance.
In each cluster, where highly similar contextual clues are grouped together, we only keep a single generated output with the highest generation probability.
The \emph{filtering} step effectively reduces potential factual errors and redundant close duplicates.
The query is then individually augmented with each filtered contextual clue.
We retrieve documents separately for every single augmented query.
%
As the last step, all the documents are ranked together~(\emph{fusion}) with the generation probability from the integral contextual clue in the augmented query.

We evaluate our approach on two established benchmarks: Natural Questions~\cite{kwiatkowski-etal-2019-natural} and TriviaQA~\cite{lee2019latent}.
Our baseline model GAR~\cite{mao2021generation} trails behind its dense retrieval counterpart DPR~\cite{karpukhin2020dense} by a large margin when retrieving a small number of passages. 
We bridge this gap and outperform GAR by 3.1\% and 2.9\% on Top-5/Top-20 accuracy on the NQ dataset. 
Compared with DPR, our approach outperforms it by 0.6 and 1.0 points on Top-100 accuracy on the two datasets, while requiring 96\% less index storage space.
The accuracy can be further improved by 3.4\% on the basis of DPR's performance when fusing the documents retrieved from DPR and our method together.
Furthermore, our retrieval performance also successfully transfers to downstream question answering tasks, where our methods increase by 3.2\% and 0.8\% Exact Match score compared with the DPR and GAR retrieved documents. 
%

\section{Methods}

\subsection{Contextual Clue Sampling and Filtering}
We employ a sequence-to-sequence model that takes the question as input and generates the \anscontext{} for the answer as target.
As our model, we use BART-large~\cite{lewis2020bart}, but note that the generator can be replaced with any other sequence-to-sequence model.
Contextual clues are the sentences in a passage that contains the ground-truth answer to the question.
These sentences are either extracted from the passage provided by the dataset~(when available), or from the matching passage used as a reference for the retriever.

At inference time, we first sample a diverse set of \anscontext{} from the fine-tuned model.
Generally speaking, a single \anscontextsingle{} can be broken into two main components; relational facts~(``august 21, 2018'') and contextual description~( ``the game was released on'').
Interestingly, we notice that many generations are identical in contextual descriptions, but inconsistent with the fact words~(various dates, numbers or named entities).
Previous works try to solve this inconsistency issue with an additional training loss~\cite{elazar2021measuring}, adding a reasoning module~\cite{nye2021improving}, or by majority vote~\cite{wang2022self}.
Instead, we first cluster the \anscontext{} based on their edit distance. 
For most cases, contextual clues with the same contextual descriptions but varying relational facts, are grouped together in the same cluster.
We then employ a simple filtering strategy for each cluster by keeping the top-ranked output with maximum generation probability, while discarding the rest outputs in the cluster.
As a result, we could gather all possible reasoning paths to the answer, while reducing potential factual errors.
As shown in Appendix~\ref{appendix:filter_effective}, the filtering strategy is crucial for the following retrieval step in terms of both retrieval \textit{efficiency}~(saves 70\% for the retrieval process) and \textit{accuracy}~(consistently better than using full contextual clues).

\subsection{Retrieval and Fusion}\label{sec:retrieval}
Defining the $n$ generated and filtered contextual clues as $\{c_{i}\}_{i=1}^n$, we augment the question ${q}$ into $\{\texttt{[CLS]} q\ \texttt{[SEP]}\ c_{i}\}_{i=1}^n$ by appending each individual context to it.
Following GAR, we use BM25 as the backend for retrieval where it could be seen as a logical scoring model using a query encoder $\eta_q$ and passage encoder $\eta_d$~\citep{lin2022proposed}:
%
\begin{equation}
    s(q,d) = \phi(\eta_q(q), \eta_d(d))
\end{equation}
%
where $\phi$ is a similarity function such as dot product or L2 distance.
Note that we use $c$ to denote the generated contexts and $d$ to denote real passages in the corpus.
To aggregate the retrieval results of different augmented queries, we perform retrieval individually for each augmented query and use the likelihood $p(c_i\mid q)$ of the generated context $c_i$ as the fusion weights.
Therefore, the final retrieval score $s_f(q,d)$ for each question-passage pair is calculated as:
\begin{equation}\label{eq:fusion}
    s_f(q, d) = \sum_{i=1}^{n} p(c_i\mid q)\cdot s(\texttt{[CLS]} q\ \texttt{[SEP]}\ c_{i},d)
\end{equation}
We finally re-sort the candidates according to the fusion scores and return the top-k passages for the next stage of answer extraction.

\section{Experiment}
\subsection{Datasets}
We conduct the experiments on two widely used ODQA datasets: Natural Questions~(NQ)~\cite{kwiatkowski-etal-2019-natural} and TriviaQA~\cite{joshi2017triviaqa}. NQ consists of 79,168 train, 8,757 dev, and 3,610 test question-answer pairs. We use the open-domain splits of TriviaQA which contains 78,785 train, 8,837 dev, and 11,313 test QA pairs~\cite{lee2019latent}.

\subsection{Experiment Setup}
We finetune the BART-large model~\cite{lewis2020bart} for \anscontextsingle{} generation. Given the question, we generate 100 candidate outputs from BART using beam search with beam size 100. We first cluster candidates using fuzzy string matching with the built-in $\textit{difflib}$ Python module. The similarity cutoff is set to 0.8 and any string pairs scoring less than the cutoff are not kept in the same group. On average, for each question, there are 24 \anscontext{} for NQ and 33 for TriviaQA after filtering. More details on processing the contextual clues are in the Appendix~\ref{appendix:filter_details} and \ref{appendix:filter_effective}.

For each contextual clue augmented query, we use the Pyserini~\citep{lin2021pyserini} BM25 to retrieve top-1000 candidate passages. All the retrieved documents are then re-ranked according to Eq.~\eqref{eq:fusion}. Further details are included in the Appendix~\ref{appendix:ret_fuse}.

\begin{table}[!t]
\centering
\resizebox{\columnwidth}{!}{
\begin{tabular}{lcccc}
\toprule
\# Context      & ROUGE-1 & ROUGE-2 & ROUGE-L & Ans Cover \\
\midrule
Top-1  & 35.27 & 22.82 & 31.84 & 29.02  \\
Full & 48.32  & 32.43   & 42.64    & 46.01   \\
Filtered & 47.14 & 31.44 & 41.80 & 43.17     \\
\bottomrule
\end{tabular}}
\caption{Evaluation of generated answer contexts on the validation set of the NQ dataset.}
\label{tab:rouge}
\end{table}

\begin{table*}[!t]
\centering
\begin{adjustbox}{width=0.9\textwidth}
\begin{tabular}{l@{\hskip 0.2in}c@{\hskip 0.2in}|c@{\hskip 0.2in}c@{\hskip 0.2in}c@{\hskip 0.3in}c@{\hskip 0.2in}c@{\hskip 0.2in}c} 
\toprule
\multirow{2}{*}{Methods} &\multirow{2}{*}{Index Size}
& \multicolumn{3}{c}{Natural Questions~}  
& \multicolumn{3}{c}{TriviaQA}       \\ 
\multicolumn{1}{c}{} & & Top-5 & Top-20 & Top-100
& Top-5 & Top-20 & Top-100   \\ 
\midrule
\textbf{Dense Retrieval}&&&&&&&\\
\quad DPR &61GB & 68.3 & 80.1 & 86.1 & 72.7 & 80.2 & 84.8 \\
\midrule
\textbf{Lexical Retrieval}&&&&&&\\
\quad BM25 &2.4GB & 43.8 & 62.9 & 78.3 & 67.7 &
77.3 & 83.9  \\
\quad GAR &2.4GB& 60.8 & 73.9 & 84.7 & 71.8 & 79.5 & 85.3 \\
\quad SEAL &8.8GB& 61.3 & 76.2 & 86.3 & - &  - & - \\
\quad Ours-single & 2.4GB& 63.0 & 75.2 & 84.8 & 71.7 & 79.1 & 84.6 \\
\quad Ours-multi &2.4GB& \textbf{63.9} & \textbf{76.8} & \textbf{86.7} & \textbf{72.3} & \textbf{80.1} & \textbf{85.8}  \\
\midrule
\textbf{Fusion Retrieval}&&&&&&&\\ 
\quad BM25+DPR &63.4GB& 69.7 & 81.2 & 88.2 &71.5 & 79.7 & 85.0 \\
\quad GAR+DPR &63.4GB& 72.3 & \textbf{83.1} & 88.9 & 75.7 & 82.2 & 86.3 \\
\quad Ours-single + DPR &63.4GB& 72.7 & 82.6 & 88.1 & 76.0 & \textbf{82.6} & 86.4\\
\quad Ours-multi + DPR &63.4GB& \textbf{72.7} & 83.0 & \textbf{89.1} & \textbf{76.1} & 82.5 & \textbf{86.4} \\

\bottomrule
\end{tabular}
\end{adjustbox}
\caption{Top-5/20/100 retrieval accuracy (\%) and index size (GB) of different models on Natural Questions and TriviaQA test sets. Each score in the right column represents the percentage of the top 20/100 retrieved passages that contain the answers. The DPR and BM25 indexes are downloaded from the Pyserini toolkit\footnotemark.}
\label{tab:main_result}
\end{table*}

\subsection{Baselines}
\paragraph{Retriever} Retrieval in open-domain QA is traditionally implemented with sparse vector space model BM25~\cite{robertson2009probabilistic}, based on exact term matching. DPR \cite{karpukhin2020dense} implements retrieval by representing questions and passages as dense vectors. GAR \cite{mao2021generation} proposes to expand the query by adding relevant answers, the title of a passage and the sentence where the answer belongs. It also fuse the results from its own and from DPR (GAR+DPR). To make a fair comparison, we extend our generation target from the answer context only (\textit{Ours-single}) to include both the answer and the passage title (\textit{Ours-multi}). We also report the fusion results with DPR.
SEAL~\cite{bevilacqua2022autoregressive}  use BART model to generate ngrams then map to full passage with FM index.

\paragraph{Reader} 
DPR \cite{karpukhin2020dense} employs an extractive reader model based on BERT~\cite{devlin2019bert} and predicts the answer span. RAG~\cite{lewis2020retrieval} combines the DPR dense retriever together with a BART answer generator, and jointly trains the two models end-to-end. 
FiD~\cite{izacard2021leveraging} also uses DPR to retrieve relevant passages and the decoder attends over all the encoded passages to generate the final answer. For fair comparison, we evaluate the retrieval results of FiD, GAR, and SEAL on the same reader model, FiD-large which takes the question along with 100 top retrieved passages as input.

\section{Results}
\subsection{Contextual Clues Evaluation}
We are interested in understanding the quality of the generated \anscontext{}. In Table \ref{tab:rouge}, \textit{Top-1} returns the top sequence with the highest probability during beam search, while \textit{Full} contexts contain all top-ranked 100 sequences. \textit{Filtered} is the set of contextual clues after filtering.
%
We report the ROUGE F-measure scores between the ground-truth and generated \anscontext{} on the NQ validation set. We also report the answer coverage rate, measured as the percentage of contextual clues that contain the answer.

As shown in Table \ref{tab:rouge}, rigorously increasing the number of generated candidates increases the ROUGE scores by at least 10\% compared with only generating the top sequence, indicating it's more probable to capture the potential ground-truth answer context. The filtering strategy effectively reduces the size of candidate contexts while maintaining high coverage and diversity (less than 1\% difference in ROUGE scores). Moreover, \textit{Full} significantly increases the answer coverage rate by $\sim 17\%$ compared with \textit{Top-1}, suggesting that not only more semantics but also more fact words are captured in a larger sizes of candidates.

\subsection{Main Retrieval Results}
In Table~\ref{tab:main_result}, we show both the retrieval accuracy and index size.
Note that the index size should be considered with salt since it largely depends on the system implementation. 
The baseline models are reported in their open-sourced versions.
We additionally compare with other memory-efficient neural retrieval models in the appendix~\ref{appendix:small_dpr_comparison} and report retrieval time latency in the appendix~\ref{appendix:latency}.

Compared with other lexical retrieval models, our method significantly outperforms both GAR and SEAL, showing the effectiveness of extensively sampled contextual clues.
We also find that \textit{Ours-multi} consistently improves over \textit{Ours-single}. 
We surmise that ground-truth answers serve as useful signals during retrieval and they are more likely to be covered when directly sampling answers.
Most of the traditional lexical retrieval methods always trail behind dense retrieval by a large margin, as illustrated in Table \ref{tab:main_result}.
Surprisingly, our method even outperforms the DPR model by 0.6 and 1.0 points in terms of top-100 accuracy on two datasets, while requiring 96\% less index storage space.
For the purpose of pushing the limit of retrieval performance, we also show the accuracy of different lexical-based methods fused with DPR.
Overall, our method fused with DPR achieves the highest accuracy across all baseline methods on both datasets.

\footnotetext{https://github.com/castorini/pyserini/blob/master/docs/prebuilt-indexes.md}

\subsection{End-to-end QA results}

As shown in Table~\ref{tab:em_score}, Ours-multi achieves the highest exact-match scores compared with other baseline methods on both datasets.
We have an interesting observation on TriviaQA dataset.
The only difference between FiD and Ours is that FiD retrieves from DPR. Although ours-single is 0.2 points lower on Top-100 accuracy than that of DPR, the EM score of Ours-single is $\sim2$ points higher than FiD. 
It demonstrates even with relatively same top-k retrieval accuracy, that our approach could retrieve qualitatively better passages that are easier for the reader model to answer.

\begin{table}
\centering
\resizebox{0.9\columnwidth}{!}{
\begin{tabular}{lccc}
\toprule
Methods     & Natural Questions & TriviaQA \\
\midrule
DPR  & 41.5  &  57.9  \\
RAG & 44.5 & 56.1 \\
FiD & 51.4 & 67.6\\
GAR & 50.6 & 70.0 \\
SEAL & 50.7 & -\\
\midrule
Ours-single & 50.6 & 69.7 \\
Ours-multi & \textbf{51.7} & \textbf{70.8} \\
\bottomrule
\end{tabular}}
\caption{
    End-to-end exact-match results on the test sets.
}
\label{tab:em_score}
\end{table}

\section{Conclusion}
We propose to narrow the lexical gap between the query and the documents by augmenting the query with extensively sampled contextual clues.
To make sure the generated contextual clues are both diverse and relevant, we propose the strategy of context filtering and retrieval fusion.
Our approach outperforms both the previous generation-based query expansion method and the dense retrieval counterpart with a much smaller index requirement.
%

\bibliography{anthology,acl, retrieval,consistency}
\bibliographystyle{acl_natbib}

\clearpage
\newpage
\appendix
\section{Appendix}
\subsection{Contextual Clues Sampling and Filtering Experimental Details}
\label{appendix:filter_details}
We finetune BART-large model \cite{lewis2020bart} for \anscontextsingle{} generation. 
For Natural Questions dataset, we extract the sentence containing the ground-truth answer from the provided positive passage. For TriviaQA dataset, since only pairs of questions and answers are provided in the original dataset, we extract the answer context sentence from the highest ranked passage retrieved by BM25. We train the model using Adam optimizer \cite{kingma2015adam} with a learning rate of $3e-5$, linear scheduling with warm-up rate 0.01, and training batch size of 256 on 4 V100 GPUs.

Given the question, we generate 100 candidate outputs from BART using beam search with beam size 100. We first group similar candidates using fuzzy string matching with the built-in $\textit{difflib}$ python module. The similarity cutoff is set to 0.8 and any string pairs scoring less than the cutoff are not kept in the same group. On average, for each question there are 24 \anscontext{} for NQ and 33 for TriviaQA after filtering.

\subsection{Effectiveness of Filtering Strategy}
\label{appendix:filter_effective}
\begin{table*}[!t]
\centering
\begin{adjustbox}{width=0.7\textwidth}
\begin{tabular}{l@{\hskip 0.2in}|c@{\hskip 0.2in}c@{\hskip 0.2in}c@{\hskip 0.3in}c@{\hskip 0.2in}c@{\hskip 0.2in}c} 
\toprule
\multirow{2}{*}{Methods}
& \multicolumn{3}{c}{Natural Questions~}  
& \multicolumn{3}{c}{TriviaQA}       \\ 
\multicolumn{1}{c}{} & Top-5 & Top-20 & Top-100
& Top-5 & Top-20 & Top-100   \\ 
\midrule
\quad Ours-single (unfiltered)  & 61.1 & 73.7 & 84.1 &70.9&78.7&84.3 \\ 
\quad Ours-single & 63.0 & 75.2 & 84.8 & 71.7 & 79.1 & 84.6 \\

\bottomrule
\end{tabular}
\end{adjustbox}
\caption{Top-5/20/100 retrieval accuracy (\%) on Natural Questions and TriviaQA test sets. Filtering strategy effectively increases the retrieval accuracy and reduces the search space for the retrieval fusion step.}
\label{tab:compare_filter}
\end{table*}

Directly augmenting question with the full set of sampled contextual clues is a sub-optimal solution due to the following reasons: 1) Retrieval efficiency: After filtering, for each query, we only need to perform 70\% less times of the retrieval. As a result, it's also saves the search space for the retrieval fusion step. 2) Retrieval accuracy: As shown in Table \ref{tab:compare_filter}, the accuracy for the unfiltered contexts is consistently lower than that on the filtered contexts. We suppose it's due to removal of hallucinated facts contained in the contextual clues during filtering.

\subsection{Retrieval Fusion Experimental Details}
\label{appendix:ret_fuse}
We put all the passages belonging to the same question but different augmentation into a public pool after filtering duplicates.
If a passage in the public pool does not appear in the top-1000 retrieval list of an augmented query, we use the minimum score of the augmented query's top-1000 list as the default score for the missing passage from that augmented query.
We then average the retrieval score for each passage in the pool according to Eq.~\eqref{eq:fusion} and re-sort the order of the fused passages.
For fair comparison with GAR \cite{mao2021generation}, we additionally fine-tune an answer generation model and a title generation model.
We perform the same fusion steps above for all three generation models, and we linearly interpolate their fusion results by searching the best weighting on the development set using Bayesian Optimization~\citep{Frazier2018ATO}.

\subsection{Comparison with Memory Efficient DPR techniques}
\label{appendix:small_dpr_comparison}
\begin{table*}[!t]
\centering
\begin{adjustbox}{width=0.6\textwidth}
\begin{tabular}{l@{\hskip 0.2in}c@{\hskip 0.2in}|c@{\hskip 0.2in}c@{\hskip 0.2in}c@{\hskip 0.2in}c} 
\toprule
\multirow{2}{*}{Methods} &\multirow{2}{*}{Index Size}
& \multicolumn{2}{c}{Natural Questions~}  
& \multicolumn{2}{c}{TriviaQA}       \\ 
\multicolumn{1}{c}{} && Top-20 & Top-100 & Top-20 & Top-100   \\ 
\midrule
\quad DPR &61GB & 80.1 & 86.1 & 80.2 & 84.8 \\
\quad DPR + PCA-256 &21GB &77.2& 85.5 &76.5&83.4\\
\quad DPR + PCA-256 + PQ &1.3GB&74.8&84.1&74.5&82.6\\
\quad BPR & 2.0GB & 77.9 & 85.7 & 77.9 & 84.5 \\
\quad DrBoost & 13.5GB & 80.9 & 87.6 & - & - \\
\midrule
\quad Ours-single & 2.4GB & 75.2 & 84.8 & 79.1 & 84.6 \\
\quad Ours-multi &2.4GB  & 76.8 & 86.7 & 80.1 & 85.8  \\

\bottomrule
\end{tabular}
\end{adjustbox}
\caption{Comparison with other memory efficient neural retrieval models on index size.}
\label{tab:index_compress}
\end{table*}

We compare our approach with other memory efficient neural retrieval models in Table~\ref{tab:index_compress}.
\citet{ma-etal-2021-simple}~show that the DPR could be furthered compressed to trade accuracy off against speed and storage.
However, the accuracy of DPR could drop significantly if compressed to the same storage level of the lexical index. 
BPR \cite{yamada2021efficient} integrates a learning-to-hash technique into DPR to represent the passage index using compact binary codes. The index size of BPR is slightly smaller than ours approach, but we achieve higher retrieval accuracy on both two datasets.
We also include DrBoost \cite{lewis2021boosted}, a dense retrieval ensemble trained in stages. DrBoost outperforms ours approach on NQ dataset, while taking $6\times$ times larger index size.

\subsection{Latency Analyses}
Retrieval time latency is an important factor to consider for deployment. We list the latency time in Table ~\ref{tab:latency}. It is notable that the latency listed in the table is tested on CPU, since we use BM25 as our retrieval backend and it only requires CPU to run. Dense retrieval methods (e.g. DPR) normally is running on GPU devices, which only takes 456.9ms \cite{lewis2021boosted} per query without other device specific speed-up techniques.

\label{appendix:latency}
\begin{table*}[!t]
\centering
\begin{adjustbox}{width=0.5\textwidth}
\begin{tabular}{l@{\hskip 0.2in}c@{\hskip 0.2in}|c@{\hskip 0.2in}c} 
\toprule
\multirow{2}{*}{Methods} &\multirow{2}{*}{Latency}
& \multicolumn{2}{c}{Natural Questions~}  \\ 
\multicolumn{1}{c}{} && Top-20 & Top-100  \\ 
\midrule
\quad DPR & 7570ms & \textbf{80.1} & \textbf{86.1}  \\
\quad DPR + PCA-256 & 2540ms &77.2& 85.5 \\
\quad DPR + PCA-256 + PQ & 765ms &74.8&84.1\\
\midrule
\quad BM25 &318ms & 62.9 & 78.3  \\
\quad GAR (ours) &962ms & 73.9 & 84.7 \\
\quad Ours-single & 1545ms & 75.2 & 84.8 \\
\quad Ours-multi &2732ms & \textbf{76.8} & \textbf{86.7}  \\

\bottomrule
\end{tabular}
\end{adjustbox}
\caption{Comparison on retrieval time latency, which is tested using 1 Intel Xeon CPU E5-2699 v4 @ 2.20GHz.}
\label{tab:latency}
\end{table*}

\end{document}